\documentclass[12pt]{article}
\usepackage{comment}
\usepackage[paperheight=11.5in, paperwidth=9.5in]{geometry}
\usepackage{lscape}
\usepackage[utf8]{inputenc}
\usepackage[final]{pdfpages}
\usepackage{amsmath}
\usepackage{bm}
\usepackage{slashbox}
\usepackage[libertine]{newtxmath}
\usepackage{color}
\usepackage{setspace}
\usepackage{pdflscape}
\usepackage{graphicx} 
\usepackage{float}
\usepackage{algorithm}
\usepackage{algorithmicx}
\usepackage{algpseudocode}	 
\pdfoutput=1

\title{ {E}volutionary Computation,  Optimization and Learning Algorithms for Data Science}
\author{Farid Ghareh Mohammadi$^1$,  M. Hadi Amini$^2$,  and Hamid R. Arabnia$^1$\\1:  Department of Computer Science,  Franklin College of Arts and Sciences, \\ University of Georgia,  Athens,  Georgia,  30601  \\
2:  School of Computing and Information Sciences, \\ College of Engineering and Computing,  \\Florida International University,  Miami,  FL 33199 \\
Emails:  farid.ghm@uga.edu,    amini@cs.fiu.edu,  hra@cs.uga.edu}
\date{}

\begin{document}

\maketitle
\section{Abstract}

{\color{red}
  }

A large number of engineering,  science and computational problems have yet to be solved in a computationally efficient way. One of the  emerging challenges is how evolving technologies grow towards autonomy and intelligent decision making. This leads to collection of large amounts of data from various sensing and measurement technologies,   e.g.,  cameras,  smart phones,  health sensors,  smart electricity meters,  and environment sensors. Hence,  it is imperative to develop efficient algorithms for generation,  analysis,  classification,  and illustration of data. Meanwhile,  data is structured purposefully through different representations,  such as large-scale networks and graphs. Therefore,  data plays a pivotal role in technologies by introducing several challenges:  \textit{how to present},  \textit{what to present},  \textit{why to present}. Researchers explored various approaches to implement a comprehensive solution to express their  results in every particular domain,  such that the solution enhances the performance and minimizes cost,  especially time complexity. In this chapter,  we focus on data science as a crucial area,  specifically focusing on a curse of dimensionality (CoD) which is due to the large amount of generated/sensed/collected data,  especially large sets of extracted features for a particular purpose. This motivates researchers to think about optimization and apply nature inspired algorithms,  such as meta-heuristic and evolutionary algorithms (EAs) to solve large-scale optimization problems. Building on the strategies of these algorithms,   researchers solve large-scale engineering and computational problems with innovative solutions. Although these algorithms look un-deterministic,  they are robust enough to reach an optimal solution. To that end,  researchers try to run their algorithms more than usually suggested,  around 20 or 30 times,  then they compute the mean of result and report only the average of 20 / 30 runs' result. This high number of runs becomes necessary because EAs,  based on their randomness initialization,  converge the best result,  which would not be correct if only relying on one specific run.  Certainly,  researchers do not adopt evolutionary algorithms unless they face a problem which is suffering from placement in local optimal solution,  rather than global optimal solution. In this chapter,  we first develop a clear and formal definition of the CoD problem,  next we focus on feature extraction techniques and categories,  then we provide a general overview of meta-heuristic algorithms,  its terminology,  and desirable properties of evolutionary algorithms.

\textbf{Keywords: } Evolutionary Algorithms,  Dimension Reduction (auto-encoder),  Data Science,  Heuristic Optimization,  Curse of Dimensionality (CoD),  Supervised Learning,  Data Analytic,  Feature Extraction,  Optimal Feature Selection,  Big Data.

\section{Introduction}

\subsection{Overview}

A large number of engineering,  science and computational problems have yet to be solved in a more computationally efficient way. One of the  emerging challenges is the evolving   technologies and how they  enhance towards autonomy. This leads to collection of large amount of data from various sensing and measurement technologies,  such as cameras,  smart phones,  health sensors,  and environment sensors. Hence,  generation,  manipulation and illustration of data grow significantly. Meanwhile,  data is structured purposefully through different representations,  such as large-scale networks and graphs. Therefore,  data plays a pivotal role in technologies by introducing several challenges:  \textit{how to present},  \textit{what to present},  \textit{why to present}. Researchers explored various approaches to implement a comprehensive solution to express their  results in every particular domain,  such that the solution enhances the performance and minimizes cost,  especially time complexity. In this chapter,  we focus on data science as a crucial area; specifically focusing on curse of dimensionality (CoD) which is due to the large amount of generated/sensed/collected data,  especially large sets of extracted features for a particular purpose. This motivates researchers to think about optimization and apply nature inspired algorithms,  such as meta-heuristic and evolutionary algorithms (EAs) to solve large-scale optimization problems. Building on the strategies of these algorithms,   researchers solve large-scale engineering and computational problems with innovative solutions. Although these algorithms look un-deterministic,  they are robust enough to reach an optimal solution. To that end,  researchers try to run their algorithms more than usually suggested,  around 20 or 30 times,  then they compute the mean of result and report only the average of 20 / 30 runs' result. This high number of runs becomes necessary because EAs,  based on their randomness initialization,  converge the best result,  which would not be correct if only relying on one specific run.  Certainly,  researchers do not adopt evolutionary algorithms unless they face a problem which is suffering from placement in local optimal solution,  rather than global optimal solution. In this chapter,  we first develop a clear and formal definition of the CoD problem,  next we focus on feature extraction techniques and categories,  then we provide a general overview of meta-heuristic algorithms,  its terminology,  and desirable properties of evolutionary algorithms.

\subsection{Motivation}
In the last twenty years,  computer usage has proliferated significantly,  and it is most likely that you could find technologies and computers almost anywhere you want to work and live. A large amount of data is being generated,  extracted and presented through a wide variety of domains,  such as business,  finance,  medicine,  social medias,  multimedia,   all kinds of networks,  and many others sources due to this spectacular growth.  This increasingly large amount of data is often referred to as Big Data. In addition,  distributed systems and networks are not performing as well as they did as in the past    \cite{Marrow2000}. Hence,  it is imperative to leverage new approaches which optimize and learn to use these devices powerfully. Moreover,  Big Data also requires that scientists propose new methods to analyze the data. Obtaining a proper result,  thus,  requires an unmanageable amount of time and resources. This problem is known as the curse of dimensionality (CoD) which is discussed in the next sub-section in detail. Ghareh mohammadi and Arabnia has discussed application of evolutionary algorithms on images,  specifically focused on image stegnalaysis  \cite{mohammadi2019isea}. But,  in this study we expanded our investigation and consider large-scale engineering and science problems carefully.

In machine learning,  the majority of problems require a fitness function which optimizes a gradient value to lead a global optimum accurately    \cite{maheswaranathan2018guided}. This function is also known as an objective function and may have different structures for different problems. In machine learning,  we work with three categories of data:  one supervised,  one semi-supervised and one unsupervised. These categories also have different learning processes based on their types. Supervised data sets are the most common data set and are characterized by having  a ground truth with which to compare results. Supervised learning algorithms normally take a supervised data sets and then divide them into two parts:  train and test. After that,  one of the supervised learning algorithms learns from train data,  predicts test data,  and compares the result with the ground truth to ascertain the accuracy of the algorithm performance. The most common types of supervised learning algorithms are classification and regression. It is noteworthy that regression has different algorithms which mainly focus on time series problems. The only exception is that regression algorithms have a particular algorithm,  Logistic regression,  which is considered as a classification,  rather than regression,  algorithm     \cite{cruyff2016review}.  In this chapter,  we focus on supervised data sets and supervised learning algorithms .

On the other hand,  unsupervised learning algorithms follow the process of using unsupervised data sets which do not have any ground truth to compare their result,  which makes classifying and evaluating the performance of the algorithm problematic. The absence of a ground truth is increasingly common through all domains such as web-based,  engineering, etc data and it is necessary to address this problem. Unsupervised learning takes more steps to analyze features and find the most relevant features with the best possible positive relation. Clustering and representation learning (RL) algorithms are the most common algorithms in unsupervised learning category. K-means is an important clustering algorithm that attempts to find k clusters located close to each other. The main problem of k-means is its bias-k towards the problem. In other words,  k-means needs to have k number set in advance before running the algorithms. RL also works for supervised data sets,  although its nature behaves in an independent way per task    \cite{zhuang2015supervised}. 

Semi-supervised data sets fall somewhere between supervised and unsupervised data sets in terms of characteristics. This means that semi-supervised learning algorithms take a data set which provides ground truth value for some instances but not for others. Expectation maximization (EM) is the most important and robust technique for working with these data sets    \cite{yang2018semi}. More over,  EM is also able to handle missing values of a given data set properly. Real data always involves missing values,  and researchers struggle with this problem.    

Feature extractor (FE) which is discussed in details in the next section,  is almost universal techniques which are capable of applying on these three types of problems to aim for dimension reduction. Meanwhile,  the majority of problems and data set have been so far used are supervised data sets. But it does not mean that FE does not apply on unsupervised or semi-supervised data sets. For instance,  for unsupervised data set,  it is normal to use dimension reduction or auto-encoder techniques for that. 

 There has been numerous challenges in the literature regarding the deployment of evolutionary algorithms for computation,  optimization and learning. These studies can be reviewed in the following major aspects:  curse of dimensionality    \cite{altman2018curse, guo2018new},  nature-inspired computation (cite all papers from 2.4 here    \cite{gupta2011reminiscent,  Marrow2000}),  nature-inspired meta-heuristic computation (cite all papers from 2.5 here    \cite{balamurugan2015stellar,  blum2003metaheuristics}),  and nature-inspired evolutionary computation (cite all papers from 2.6 here    \cite{back2018evolutionary,  pierezan2018coyote,  mohammadi2014ifab,  mohammadi2014IFABKNN,  pierezan2018coyote}).  These studies are elaborately reviewed in the following.

\subsection{Curse of Dimensionality}
Curse of dimensionality is related to the fact that the input data is too huge that no human being can analyze it. In Machine Learning,  recently,  researcher work with high-dimensional data. For instances,  if we’re analyzing 3 channel images,  such as RGB,  HSV images ,  sized 512x512,  we’re working in a space with 512*512*3 dimensions. Altman and Krzywinski    \cite{altman2018curse}  believe that having more data is much better than having few or nothing. This overabundance of data is called the curse of dimensionality (CoD) which causes problems in big data era such as  data sparsity,   multiple testing,  which researchers    \cite{guo2018new} proposed a new approach to solve the problem,  and most importantly over-fitting which is opposite of under-fitting. Beside these problems,  CoD also brings high time complexity problem which makes scientists suffering from waiting too much time to get a result. 

The world of Information technology CoD not only causes a wide range  of problems to scientists,   but also has a wide adversely affect other majors,  such as engineering    \cite{wunsch2019uncertainty},   medicine    \cite{barbour2019precision,  karpagam2019automated},    cognitive science    \cite{vong2019additional,   patel2018curse},  bioinformatic    \cite{oudah2018taxonomy},   and even optimization problems    \cite{serani2019stochastic,  gupta2019big,  maheswaranathan2018guided}.

Classification in Big Data suffers from plenty of problems and issues,  one of which is considered very challenging named CoD. Traditional feature extraction techniques also are not able to solve this problem technically any more due to some limitation   \cite{gupta2019big}. According to the research studies have accomplished,  scientist proposed a new approach to solve this problem. Researchers introduce nature-inspired computation which enable to simulate traditional feature extraction techniques in a way that improve the performance of classification. 

\subsection{Nature Inspired computation}
Pure and basic machine learning algorithms are not capable of solving emerging challenging issues in the world of technologies any more. It is needed to adopt a new approach to face this problems and leverage decent machine learning algorithms. Finally,  scientists discovered that combining machine learning algorithms in a technical way may solve the problems. This mixture of machine learning techniques is called nature-inspired computation,  but it still is considered an advanced machine learning algorithms.

Majority of scientific and technological developments leverage inspiring from the nature towards their goal,  especially robotics simulate how the nature works. In world of computer science,  each tool or software development process is needed to have strong synchronization,  robustness,  manageability,  parallelization,  scalability,  distributedness,  redundancy ,   adaptability,  cooperation. Indeed,  the nature provides the same properties. Therefore,  the nature-inspired techniques play an important role in computing environments.  Concretely,  the nature-inspired techniques are adopted to develop practical algorithms to solve data-driven optimization problems   \cite{gupta2011reminiscent}. 

Researchers in    \cite{Marrow2000,  gupta2011reminiscent}  categorized nature-inspired computation. In    \cite{gupta2011reminiscent} authors classified them into six different categories such as swarm intelligence,  natural evolution,  molecular biology,  immune system  and biological cells . But here,  we provide another applicable way to express the nature-inspired computation towards solving problems. One meta-heuristic and one evolutionary computation.

\subsection{Nature-inspired Meta-heuristic computation }

A meta-heuristic is an advanced procedure developed to seek and generate a sufficiently tuned solution to data-driven optimization problems.   \cite{balamurugan2015stellar}. It involves , high level view,  two types of computations. The first and foremost one is population based computation which is well-known as an evolutionary algorithms,  second one is non-population computation such as Tabu search (TS),  stochastic local search (SLS),  iterated local search (ILS),  guided local search (GLS). For more information about this classification, please refer to  \cite{blum2003metaheuristics}. Further,  Razavi and Sajedi    \cite{razavi2018svsa} proposed a single-based meta-heuristic algorithm,  Vortex Search Algorithm (VSA),  is inspired by the vortices.  In this chapter,  we mainly focus on the former classification,  evolutionary algorithms which is discussed next sub-section properly.
\subsection{Nature-inspired evolutionary computation }

Evolutionary algorithms (EAs) is invented not more than 28 years and is not pretty old computational algorithm    \cite{back2018evolutionary}.   Research studies have been accomplished new evolutionary algorithms in engineering and computational science    \cite{pierezan2018coyote, mohammadi2014ifab,  mohammadi2014IFABKNN}. EAs are known as population based algorithm. Their learning process comes from interactions between multiple candidate solutions called food source or population. EAs are particular optimization type of  meta-heuristics designed to solve optimization problems    \cite{pierezan2018coyote}.  This chapter discuss classical EAs and  other popular methods including memetic algorithms (MA),  particle swarm optimization (PSO),  and artificial bee colony (ABC),  ant colony optimization (ACO),   grey wolf optimizer (GWO) and coyote optimization algorithm (COA).

\subsubsection{ Evolutionary-based Memetic algorithms}
Memetic algorithms (MAs) are one of particular growing research studies within EA. Based on a population based search and local search,  MAs have practically succeeded in a variety of engineering and science problem domains,  in particular for NP-hard optimization problems    \cite{moscato2019accelerated, pierezan2018coyote}. Memetic algorithms intrinsically exploit all available sources,  however,  traditional EAs fail to do that. Population based search MAs leverage recombination (or crossover operator) which is an important process within MAs.For the search process,  it is essential to have three parameters ready:  one neighborhood relation,  one guiding function,  and a search space which provides borders of the problem.

The search space is also important to provide comprehensive knowledge for guiding function works. The implication of search space is to influence the dynamics of the search algorithm. These dynamics stand for the relationships,  which are accessible,  among the configurations. Thus,  these relationships depend on neighborhood function. For more information about this topic,  please refer to   \cite{moscato2019accelerated}.

\subsection{Organization}

\begin{figure}[H]
    \centering
    \includegraphics[height=3.9in]{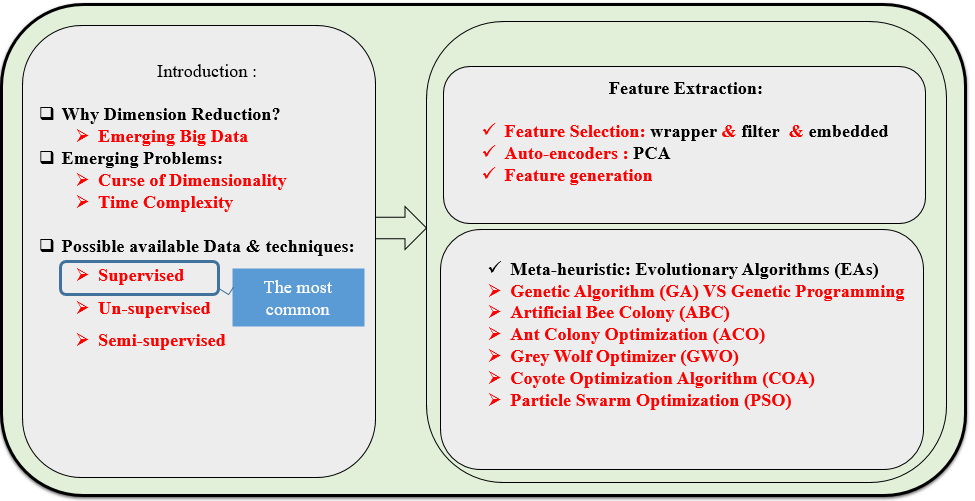}
    \caption{Overall structure of this study}
    \label{fig: General_Structure}
\end{figure}

The rest of this study is organized as follows. In section 2,  we have discussed the feature extraction techniques and their categories. First,  feature extraction from a sample object like image against feature extraction from given data sets are mentioned. Next,  the feature extraction from data set has selected to discover it. It has three types including feature selection,  auto-encoder and feature generation. Then,  we introduce nature-inspired algorithms and their application,  together with related pseudocode in solving  large-scale engineering and science problems,  particularly CoD problem. The summary of evolutionary algorithms have been discussed in this chapter is as follows:  genetic algorithm (GA),  artificial bee colony (ABC),  ant colony optimization (ACO),  grey wolf optimizer (GWO),  coyote optimization algorithm (COA) and particle swarm optimization. In general,  Figure \ref{fig: General_Structure} represents the overall structure of this study.

\section{Feature Extraction Techniques}
It is worth mentioning,  in the world of science,  "feature extraction" is used to refer to two completely separate applications. They are two different processes,  one occurring before raw data generation and one taking place after data has generated. The process of feature extraction before having raw data works to extract features using some advancing techniques,  to export information from the objects. For example,  if we want to extract features from images,  we need to adopt advanced image processing techniques,  like a feature extractor,  for that end.  Therefore,  based on the generated data,  we will have a set of raw data. Then,  in pre-processing techniques,  a second type of feature extraction is used for dimension reduction. Three major differences separate these two types of feature extraction. The first difference is their input value; the input value of the first algorithm is not particular features,  but the second feature extraction accepts only features of any data set. Second,  the first type of feature extraction is domain specific,  while the second type is not domain specific. Third,  the former does not adopt machine learning algorithms,  but the latter type does. Basically,  both of them work with data,  take values and generate outputs. The scope of the first algorithm is dynamic and would be any multimedia or social networks,  etc. On the other hand,  the second one has a almost stationary scope of input data. 

General overview of testing and evaluating given data set is shown in fig \ref{fig: EV-general }. On the top of the figure,  it clearly presents that three separate steps are required to be done in advance before generating a proper result. Pre-processing plays a main role in each problem of engineering and optimization problems. Then,  A classification algorithm is selected to make a model based on the train data. finally,  the classifier attempts to predict the test data based on the learned data.

Once data is generated and data set is ready to be evaluated,  we call the data set,  raw data set. This data set is needed to be converted into a standard data set which enables classifiers to examine in a professional way and obtain a higher performance. The most common problems of raw data set consist of curse of dimensionality (CoD),  heterogeneous features in case of values and type,  missing values,   outliers. In this chapter,  we discuss in detail how evolutionary algorithms (EAs) are adopted to solve the CoD problems,  the bottom of the figure \ref{fig: EV-general } depicts the idea where EAs are explicitly embedded into pre-processing and enhances the classifier's performance. Concretely,  EAs attempts to optimize the process of feature extraction in an innovative way.

Feature extraction (FE),  which is one of the most popular pre-processing techniques,  is the process of shrinking the number of dimension (features) and the capability of having adapted diversity while considering strong mapping between features and target values. FE aims to decrease the feature dimension as minimum possible as it keeps the same performance. A feature extractor is considered as the best one which is capable of decreasing the feature dimension and meanwhile improving the performance. The better result obtained by the better FE. FE  techniques are intrinsically classified into three broad groups:  one auto-encoder,  one feature selection (FS) and feature generation. The first two of which are the most common techniques in the scope of dimension reduction. Meanwhile researcher can leverage feature generation ( such as    \cite{shi2018efficient,  khurana2018feature} )to improve a classifier performance. The former technique is also known as dimension reduction (DR) which attempts to transform given dimension to a new dimension with  strong linear connectivity of original dimension. The most popular auto-encoder algorithm is principal component analysis (PCA). 

PCA completely is used to generate a new dimension using a certain formula and convert the given data into new dimension. The idea behind PCA is that it leverages singular value decomposition (SVD) theorem to seek for the most relevant and correlated features and the relationship between each others.  Although PCA is used to emphasize variation and bring out strong patterns in a data set,  it may not guarantee to reach a optimal solution in some data sets. PCA fails once your special visualization of instances which leads to loss of information. It tries to convert input data into new dimension using a linear function.  Circle-based and sine or cosine-based distribution of instances are the most popular situations that PCA fails.  PCA failure means that the FE did not obtain a better performance while decreasing the dimension,  not only that,  but also it did not yield the same performance. If PCA does not yield a better result,  it means that features are not correlated or have non-linear relationships.  However,   researchers often used to enable data easy to explore and visualize,  in case for representation learning (RL)    \cite{zhang2018network}.

Feature selection (FS),  the latter one,  which is the process of choosing proper sets of relevant features rather than converting to a new dimension. FS covers the lack of auto-encoders properly by keeping the original values of features during the process,  meanwhile it is most likely to decrease the number of features / dimension. Feature selection mainly provides three kinds of categories:  filter-based,  wrapper-based and embedded FS. Filter-based FS is the easy technique to implement and can be adapted to each engineering problems independently. It tries to examine given data set features separately non-dependently with respect to their target. It attempts to calculate the goodness of each feature separately. However,  the wrapper-based feature selection relies on a set of selected features and calculated their goodness using classifiers. Wrapper-based FS is a special kind of filter-based FS such that wrapper-based FS has capability of using some hyper-parameter function for evaluation. Therefore,  the pace of running filter-based is high in comparison with wrapper-based. So,  it is recommended for real-time systems because of low time complexity. Furthermore,  filter-based is cheaper than wrapper-based. But the wrapper-based feature selection    \cite{mohammadi2014ifab,  vanaja2018novel,  hancer2018pareto} yields a better result than filter-based feature selection. By advancing technologies,  wrapper-based FS also can be adopted in every system,  even real-time decision making system    \cite{vanaja2018novel}. The third one,  embedded feature selection which is similar to the wrapper-based feature selection to select the best subsets of features. However,  it has a important drawback,  which is time complexity in comparison with earlier feature selection,  when it tries to train the model. One of the popular embedded feature selections is regularization which provides both training and making model section,  together with automatic feature selection at the same time. Furthermore,  researchers    \cite{liu2018hybrid,  rostami2016PSO},  proposed another type of feature selection,  combined (hybrid) methods,  which mixes evolutionary algorithms together with filter based or wrapper based algorithms. 

Feature generation,  is considered the third type of feature extractor techniques. Feature generation is a technique between feature selection and dimension reduction. It starts to examine the features and tries to generate features using the features. In this case,  you first increase the feature dimension then remove irrelevant features. Unlike dimension reduction,  no new dimension is generated.Feature generation keeps the original features for generating new features. Then,  Feature generation can do feature selection based on the generated features    \cite{shi2018efficient}. 
\section{Bio-inspired evolutionary computation}
Engineering problems and other sensitive optimization need to reach the global optimum. However,  machine learning algorithms are not useful anymore. So,  it is required scientists adopt new kind of algorithms have been proved completely in nature for years. In this section,  we provide general overview of nature-inspired algorithms and their terminology. Tables \ref{tab: AbbreTable} and \ref{tab: AbbreTable1} provide complete definitions for abbreviation which are used in this chapter.  

\begin{table}[H]
    \centering
    \vline
    \begin{tabular}{c|l|}
     \hline
    Abb & Definition\\
     \hline
     ABC & Artificial bee colony \\
     ACOAR & Ant colony optimization attribute reduction \\
     BA & Bee algorithm \\
      
      BCO & Bee colony optimization \\
      BOA & Butterfly optimization algorithm \\
      CNN & Convolutional neural network \\
      COA & Coyote Optimization Algorithm \\
      CoD & Curse of dimensionality\\
      CSO & Chicken swarm optimization\\
      CCSO & chaotic chicken swarm optimization \\
      CRO & Coral reefs optimization\\
      DA & Dragonfly algorithm \\
      DR & Dimension reduction \\
      EAs & Evolutionary algorithms\\
      FE & Feature extraction \\
      EM & Expectation maximization \\
      EP & Evolutionary programming \\
      FS   & Feature selection\\
      FSA & Fish swarm algorithm\\
             \hline
            \end{tabular}
  
    \caption{List of Abbreviations}
    \label{tab: AbbreTable}
\end{table}

\begin{table}[H]
    \centering
    \vline
    \begin{tabular}{c|l|}
     \hline
    Abb & Definition\\
     \hline
      GA & Genetic algorithm \\ 
      GANs & Generative adversarial networks \\
       GGA & Generational genetic algorithm \\
      GLS & Guided local search \\
      GP & Genetic programming\\
      GWO & Grey Wolf Optimizer \\
      HBMO & Honey bee mating optimization \\ 
      IFAB &   Image steganalysis using FS based on ABC \\
      IoT & Internet of things \\
      ILS & Iterated local search \\
      IWOA & Improved whale optimization algorithm \\
      MAs   & Memetic algorithms\\
      ML   & Machine learning \\
      PCA  & Principal component analysis \\
      PEAs & Parallel evolutionary algorithms\\
      RFPSO & RelieF  and PSO  algorithms  \\
      RL   & Representation learning\\
      RNN & Recurrent neural network \\
      
      SLS & Stochastic local search \\
      SSGA & Steady state genetic algorithm  \\
      SVD & Singular value dimension \\
      SVM & Support vector machine\\
      TMABC-FS & Two-archive multi-objective ABC algorithm for FS\\ 
      TS & Tabu search \\
      VSA & Vortex Search Algorithm \\
      WOA & Whale Optimization Algorithm\\
      WANFIS &Whale adaptive neuro-fuzzy inference system\\
       \hline
            \end{tabular}
  
    \caption{List of Abbreviations (Continued)}
    \label{tab: AbbreTable1}
\end{table}

\subsection{Overview of evolutionary algorithms}
\begin{figure}[H]
    \centering
    \includegraphics[height=2.8in]{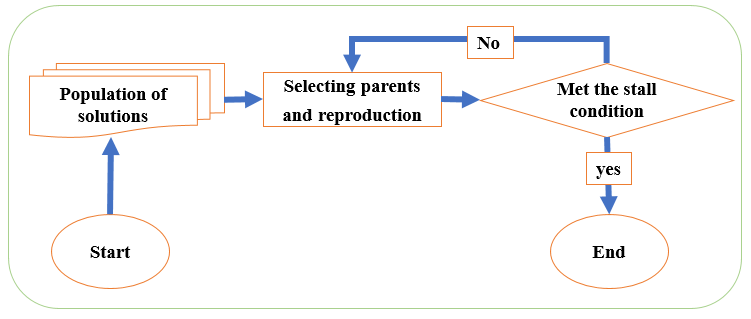}
    \caption{General process of evolutionary algorithms }
    \label{fig: EV-general }
\end{figure}

 Everything in EA starts to explain the problem and proper solutions. The first important step in evolutionary algorithm is representation. After that,  in each step,  EA works based on this representation. Figure \ref{fig: EV-general } depicts a general overview of each evolutionary algorithm's  procedure. It is extremely necessary how to present your sample solutions. Two approaches are given:  an one-hot representation and an integer representation. The former one is also known as binary representation. The number of "1" in the solution shows the number of parameters have to be involved to yield a result."1" represents that which specific features are selected and "0" stands for the features which are not considered in a specific solution. In this case,  your solution's length would be as same as the input feature dimension. If feature dimension become too big,  handling the food source are going to be a challenging issues which waste resources and yields high time complexity. However,  the integer representation works good even with high feature dimension. But it still has a big disadvantage which you need to set the reduced length of your feature vector in initialization step.
 
 Second important step is generating a population based on the descriptive model of representation. This population mostly is generated randomly with considering the representation limitation. Third step is fitness function and evaluation process. It is important to provide a tuned fitness function (objective function) towards their application of the evolutionary algorithms.
 
 The next step is to select two possible solutions as parents of new generations. Selection strategy has two broad categories. One uniform parent selection and one un-uniform parent selection. In the former one,  each solution has the same chance to be selected. However,  the latter one has different structures and criteria,  and parents are selected based on those. The un-uniform parent selection has different strategies,  the most important strategies is proportional selection which is also as known as roulette wheel,  ranked based selection,  and tournament selection.
 
 Roulette Wheel and Tournament are the most widely used selection methods in GA.
 Roulette consider the fitness value fore each chromosomes with respect to their probabilities,  using the equation \ref{eq.RoletteWheel} where p[i] stands for the probability of selecting a specific chromosome i,  f[i] goes for the fitness value of each chromosome of index i.
 
 \begin{equation}
p[i]= \frac{f[i]}{\sum f[i]} 
 \label{eq.RoletteWheel}
\end{equation}

Moreover,  the Tournament selection is pretty simpler than Roulette wheel. The idea is that it takes k chromosomes and selects based on the fitness value of each chromosome.  The best fitness value goes for the lucky chromosome to be selected.

 After that,  EA tries to reproduce new generation and updated the population. EA take two parents and regenerates new offspring based on crossover operator. The crossover or recombination,  which is one of genetic operators used to recombine two chromosomes to generate new offspring. The crossover operator includes uniform crossover,  arithmetic crossover and  k-point crossover which is a classical one. Once crossover step is done,  mutation should be done with a specific rate. The mutation may change one or more components.
 
 Finally,  the stall condition which is set to check once new generation produced. If the new generation met the condition,  EA stops running and return the best the solution which satisfied the condition.
\subsection{Genetic algorithm v.s genetic programming }
It is a common mistake that to think Genetic algorithm (GA) is the same genetic programming (GP). Generally speaking,  researchers have used these two algorithms interchangeably. But,  from a technical point of view they are completely different techniques. In this sub-section,  we provide a clear definition of each of them.
\subsubsection{Genetic algorithm}
Genetic algorithm is one of the basic but important evolutionary algorithm. It has been applied on majority of problems such as engineering,  medicine,  finance,  etc.  GA provides two kinds of approaches towards solving problems    \cite{jiang2016steady}. One steady state genetic algorithm (SSGA) and one generational genetic algorithm (GGA). They are different based on their procedure and updating mechanism function of whole process,  but they do the same process of parent selection,  reproduction and population update. In the literature,  some studies deployed GA as an effective tool for solving large-scale optimization problems,  including optimal allocation of electric vehicle charging station and distributed renewable resource in power distribution networks    \cite{PSO2017app},  resource optimization in construction projects    \cite{GA2001app},  and allocation of electric vehicle parking lots in smart grids    \cite{GA2014app}.   Algorithm \ref{tab: GA}    illustrates a pseudo code of basic GA in detail.

 \begin{algorithm}[H]
	\caption{Implementation of GA algorithm for feature selection}
	\begin{algorithmic}[1]
		\Require  $S=\{x_0,  x_1, x_2, ..., x_n\}$, $ max_{iteration} \ge 0$,  t=0,  $\alpha_M \in [$0, 1$]$,  $random_{number} \in [$0, 1$]$,  $Best_{solution} = \emptyset$.
		\Ensure $Best_{solution} :  An optimal subset of features ($F$)$ , $F=\{x_0, x_1, x_2, ..., x_m\}$ ,   m$\le$ n , $({\forall f_i \in F})\in S$ ,  $ F_{length} \le  S_{length}$.
	    
	    \For {t=0 $\cdots$ $max_{iteration}$}
				\State  Call parent selection function
		\State Call crossover method to generate offspring
		 \If{$random_{number} \le \alpha_M$}
	     \State{ Call mutation function}
		 \State	{Return offspring}
	\EndIf
		\State Call fitness function to evaluate the chromosome 
	   \If{ any chromosome obtained the best score}
	   \State \hspace{3mm} {Update the $Best_{solution}$ }
	   \EndIf
	   
		\EndFor
	\end{algorithmic}
	\label{tab: GA}
\end{algorithm}

\begin{figure}[H]
    \centering
    \includegraphics[height=3.3in]{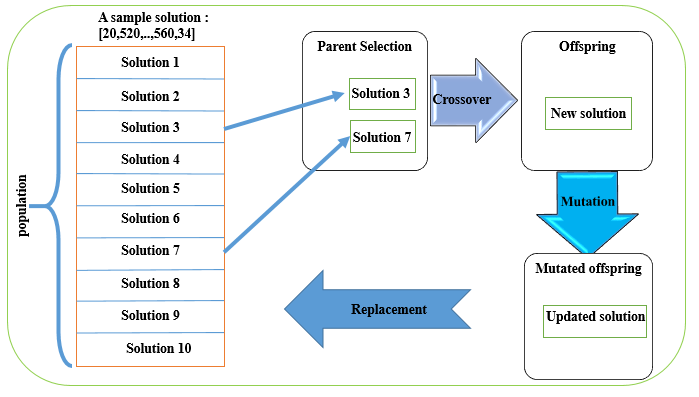}
    \caption{SSGA (steady state genetic algorithm):  Process of updating the population}
    \label{fig: SSGA}
\end{figure}

In SSGA,  GA works with a stationary population which the size of that will be the same and just it's solutions get updated each iteration. Moreover,  SSGA is an in-place algorithms which their population do not need another space to update. Like normal process,  SSGA also starts with a problem representation and fitness function,  then initialize the selection strategy,  crossover and mutation operators. After that,  SSGA takes another step to update the population with replacement strategy. Figure \ref{fig: SSGA} depicts that how two solution are selected,  crossover and mutation operators are applied and then new solution is replaced with the worse solution.

\begin{figure}[H]
    \centering
    \includegraphics[height=3in]{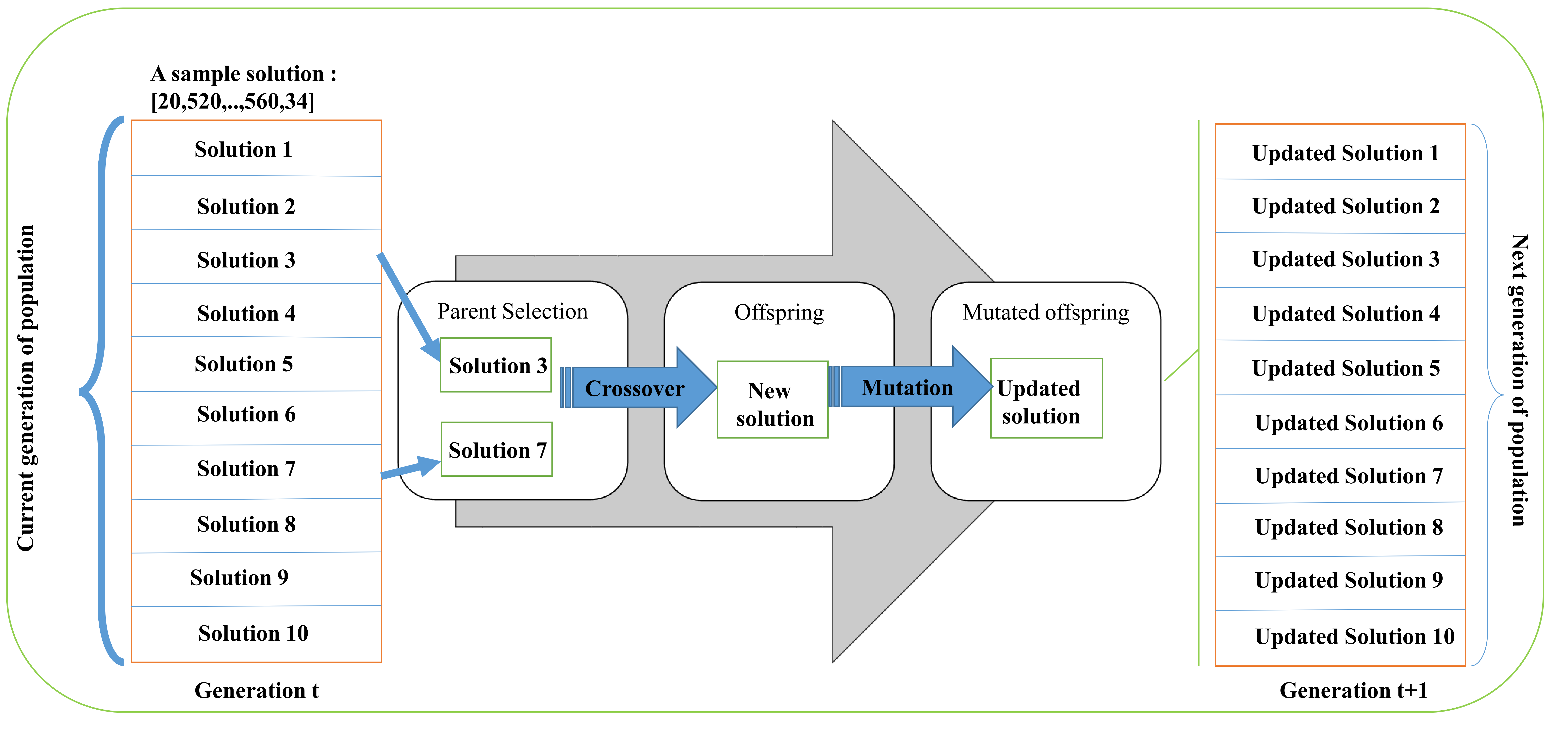}
    \caption{GGA (generational genetic algorithm):  The process of generating a new population (generation t+1)}
    \label{fig: GGA}
\end{figure}

Further,  GGA produces a new population each iteration. So,  GGA is not a in-place algorithm since it generates a new population each iteration. GGA follows the same structure of EA except the last step which is replacement. GGA skips this step since it generates a new population in each iteration. Therefore,  the replacement step is not required. Figure \ref{fig: GGA} shows complete process of generating a new population (generation t+1) from current population (generation t).

From technical point of view,  scientists can apply either GGA or SSGA based on the problem model and strategies. However,  SSGA converges faster than GGA since parents always are selected through the same population and then replaced the worse solution with the another best solution. Hence,  most of research studies are accomplished using SSA. Moreover,  most Evolutionary algorithms are discussed here also use the same strategies to converge faster towards global optimum. But,  SSGA still has a disadvantage that may stuck in a local optimum. 

\subsubsection{Genetic programming}
\begin{figure}[H]
    \centering
    \includegraphics[height=3.4in]{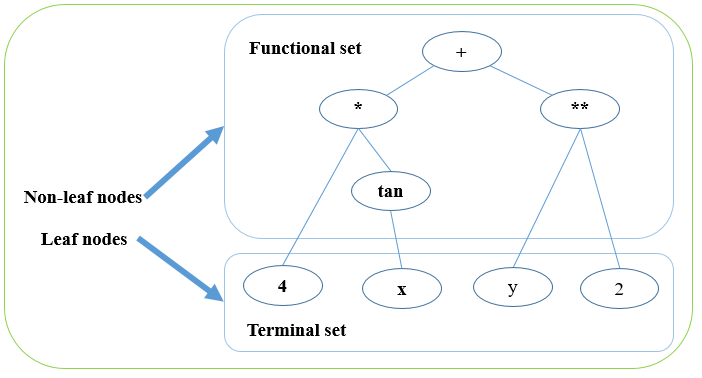}
        \caption{Tree  presentation of a problem}
    \label{fig: GP-example }
\end{figure}

Genetic programming (GP) is proposed by Koza in 1992    \cite{koza1992genetic}. It is noteworthy that this idea is introduced date back to 50s. GP evolves computer programs which are represented as trees. Each tree consists of two sections:  a function set and second is terminal set. Both of them provides constant sets of symbols. The former one always play non-leaf nodes role and the latter one plays leaf nodes role. Figure \ref{fig: GP-example } shows an example of presenting a problem \begin {math}4*tan(x)+ y^2.\end{math}.

\begin{figure}[H]
    \centering
    \includegraphics[height=3.5in]{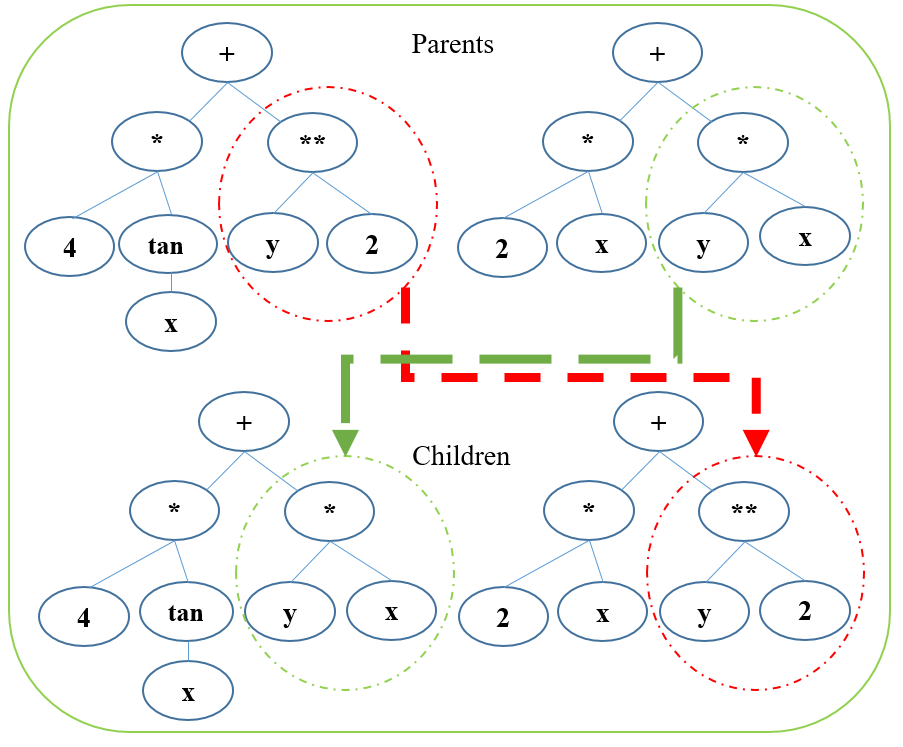}
    \caption{Crossover operator in genetic programming}
    \label{fig: GP_crossOver}
\end{figure}

Similar to GA that crossover is conducted on vectors,  in GP crossover is done through a tree and only needs to choose two sub-tree. Figure\ref{fig: GP_crossOver} expresses that the first two tree has two subset which are selected as a parent. Second tree the below are the new offsprings which are generated based on parents. GP is mostly generational genetic algorithm. Thus,  GP is not a in-place algorithm. GP is useful for solving engineering and computational problems (e.g.,        \cite{russo2018knowledge}).

Genetic programming has specific advantages over genetic algorithms. Here,  we address the most important characteristics of GP. Genetic programming has a wide variety of representation model which makes it pretty flexible against genetic algorithm. This flexibility of GP comes from it's tree-based properties. Another important feature of GP is its application over GA. GP has greater applications in comparison with GA. In spite of considering positive features of GP,  it also has disadvantages which should bear in mind. The most disadvantages of GP is its speed which is extraordinary slow. Another point is its lack of handling a large number of input data which makes also hard to handle required related population.

There is still another algorithm that attracts researcher's attention called evolutionary programming (EP). Fogel \emph{et al}    \cite{fogel1965artificial}
originally introduced evolutionary programming. It is classified as one of the major evolutionary algorithms. It resembles genetic programming,  but it does have a non-variable structure of the program to be optimized. Classical EP develops gradually finite state machine or every structure similar to it. EO always works with mutation only and does not consider crossover at all. It worth mentioning that EP uses a fitness function based on the training sequences. This feature enables EP yields a better result for prediction in time series problem and sequence problems like DNA and RNA.

\subsection{Artificial bee colony algorithm}

In the bees population,  the process of mating and generating new offspring,  finding new food sources and gathering the nectar,  sharing information in hive,  allocating tasks,  onlooker and scout bees; all of these have been inspired properly and nature-based evolutionary algorithms have been presented. To be specific about the algorithms,  honey bee mating optimization (HBMO),  bee colony optimization,  bee algorithm (BA) and artificial bee colony (ABC) are the most popular research studies are accomplished based on these algorithms    \cite{karaboga2014comprehensive}. Karaboga  \emph{ et al}    \cite{karaboga2014comprehensive} presents statistical overview of using these algorithms in scientific papers. It is worth mentioning that ABC has received the highest amount of usage with respect to the its application in engineering and science problems. Among all research studies had been done,  according to the    \cite{karaboga2014comprehensive} ABC,  BA,  BCO and HBMO are found the most useful application,  from the highest number to the lowest number,  in large scale engineering problems. ABC has been considered as the most useful algorithm in several different fields and majority of research studies leverage ABC in their problems,  such as:  training neural network (NN),  solving electrical,  mechanical, software,  control and civil engineering problems,  facing wireless sensor networks issues,  optimizing protein structure and most importantly solving image processing problems. In this chapter,  we address emerging challenges like CoD problem in Big Data and provide practical engineering solutions using ABC and other related algorithms.

Here,  we will discuss artificial bee colony (ABC) which is inspired by a set of sequential processes such as the process of seeking for a bunch of flowers,  sharing information in the hive regarding that and allocating employed,  onlooker and scout bees. Karaboga introduced ABC    \cite{karaboga2005idea} which is compatible with continues  problems in 2005. Algorithm \ref{tab: ABC_general} presents a general procedure of given ABC. A large number of research studies have accomplished using this algorithm    \cite{zabihi2018novel,  cao2018improved} and even convert that into a way that it also works with discrete problems    \cite{mohammadi2014ifab,  hancer2018pareto,  karpagam2019automated}. Not only those,  but also ABC is applied on optimization problems as an optimizer    \cite{xue2018self,  harfouchi2018modified,  wang2018improved,  cao2018improved}

Artificial bee colony interact with three groups of bees to have work done. The first group is employed bees,  second is onlooker and last one is scout. In initialization step the number of these group are set. The employed bees,  together with onlooker bees create a population which has an equal amount of two groups. ABC starts with initialization step which has positive impact on converging in ABC. Among initialization variables,  limit is important criteria and provides a condition when an employed bee converts into scout bee; at a time,  we only have one scout bee.

 \begin{algorithm}[H]
	\caption{Implementation of ABC algorithm for feature selection }
	\begin{algorithmic}[1]
   
		\Require  $S=\{x_0,  x_1, x_2, ..., x_n\}$,  $ P_{size} $=$ 2*n $,  $limit  \ge 0 $,  $ 0 \le  lower_{Bound} \le n/2$,   $ lower_{Bound} \le  upper_{Bound}  \le n$, $ max_{iteration} \ge 0$,   t=0 , ${ v}$= $random_{number}  \in  [$0, 1$]$ , $ {v'} $= $ random_{number}   \in  [$0, 1$]$,  $Best_{solution} = \emptyset$.
		
		\Ensure $Best_{solution} :  An\ optimal\ subset\ of\ features\ ($F$)$ , $F=\{x_0, x_1, x_2, ..., x_m\}$ ,   m$\le$ n , $({\forall f_i \in F})\in S$ ,  $ F_{length} \le  S_{length}$.
    	
    	\State {Call fitness function to evaluate the whole food source (S)  (primary evaluation of each food)}
		\For \{t=0 $\cdots$ $max_{iteration}$\}
		
		\State Call Employed bees to update the food source regarding their evaluation
		\State Call Onlooker bees to exploit the local foods to generate new food (solution)
		 \State Choose parents and generate  a new food (solution) based on  ${V_i}$= ${f_i}$+${v * (f_i-f_j)}$
		 \If{limit is met} : 
	     	\State \{ Call scout bee to explore new (unseen) food source to prevent from local optimum using 
	     	\State${X_i}$= ${X_{upper_{Bound}}}$+${v'}$ * \{ (${X_{upper_{Bound}}}$-${X_{lower_{Bound}}}$)\} \}
	     	
		 	\Return ${New Solution}$
	\EndIf
		\State Call fitness function to evaluate the Solution 
	   \If{ any Solution obtained the best score}
	   \State \hspace{3mm} \{Update the $Best_{solution}$\}
	   \EndIf
	   		\EndFor
	\end{algorithmic}
	\label{tab: ABC_general}
\end{algorithm}

\subsection{Particle swarm optimization algorithm}

The particle swarm optimization (PSO) is one of the  population-based meta-heuristic algorithms and optimization techniques. PSO is inspired from social–psychological principles    \cite{chen2019hybrid}. In 1995  Particle swarm optimization first introduced by Kennedy and Eberhart    \cite{kennedy1995eberhartPSO}. The PSO is based on the simulation of common animal social behaviors,  for instances:  fish schooling,  bird flocking. PSO like other evolutionary algorithms searches for the global optimum rather than local optimum. However,  the particle swarm trapped into local optimum easily when feature dimension grows significantly. Algorithm \ref{tab: PSO_general} presents a pseudo code for a standard PSO. The whole process of PSO usually initialize groups of random particles and computes fitness for each particle within iterations in order to converge into global optimum. Each particle is considered as a single solution to our problem. 

PSO follows two simple yet essential steps to have completed optimization process to find the minimum optimum or maximum optimum. The first step is communication among particles. Each particle shares their information with other particles after moving in their direction. This process makes them find a proper way toward the goal. Each time,  based on the problem (maximum / minimum optimization),  particles follows the particle and consider the particle that match the problem goal. For instance,  each iteration particles call fitness function to get fitness of their location. Then,  among the particles,  one has the best value which is set to best personal location. The best value is examined based on the problem,  if it is minimum optimization then the best value goes for the particle that has the minimum value. Moreover,  if the problem is maximum optimization the best value goes for the particle that has the maximum value. When this value is set,  each particle updates their direction and moves toward this values. It is obvious that the one has the best value does not move unless other particles find the best value. The second step which each particle does is to learn. They can learn how to update their direction after each iteration and tune the parameters.

The PSO does not have  parent selection,  recombination and mutation steps   \cite{kennedy2010particle}; thus,  this enables PSO to behave in a particular way in comparison with other evolutionary algorithms.  Concretely,  each member within the population do not get updated nor removed. Hao \emph{et al}    \cite{hao2007particle} introduced a new PSO with added crossover operator. Zhang \emph{et al}    \cite{zhang2014binary} proposed a binary PSO with mutation operator to address CoD problem using feature selection techniques to solve it. The crossover enables the particles  does not stop in the local optimum by sharing the other particles' information.  In     \cite{gupta2019big} PSO is classified into three different versions:  classical PSO,  scale-free PSO and binary PSO. 

Few parameters are required to adjust,  and enable PSO easy to implement,  make popular  stochastic and yet powerful swarm-based algorithm. Inertia weight becomes more important than other due to it's ability of having a trade-off between the exploration and exploitation process within a search space. In addition,  inertia weight has positive affect convergence rate in PSO     \cite{agrawal2019particle}.

\begin{algorithm}[H]
	\caption{Implementation of PSO algorithm for feature selection }
	\begin{algorithmic}[1]
   
		\Require  $S=\{x_0,  x_1, x_2, ..., x_n\}$ ,  $ particles_{number}  \ge 1$,   $ acceleration_coefficients (c_1 ,  c_2)  \in [$0, 1$]$,  $ max_{velocity}$ ,   t=0, ${min_weight,  max_weight}=random_{number} \in  [$0, 1$]$ ,  $ Best_{solution} = \emptyset$ .
		
	       \Ensure $Best_{solution} :  An optimal subset of features ($F$)$ , $F=\{x_0, x_1, x_2, ..., x_m\}$ ,   m$\le$ n , $({\forall f_i \in F})\in S$ ,  $ F_{length} \le  S_{length}$.	
    	 
    	 \For {t=0 $\cdots$ $max_{iteration}$}
    		\For {i=0 $\cdots$  $particles_{number}$}
		\State {Call fitness (= objective) function to for the current particle}
		\State Save the best personal location
		\State Save the best global location 
		\EndFor
		 \State  Update the ${inertia}_{weight}$
		  \For {i=0 $\cdots$ $particles_{number}$}
	     	\State Update the velocity
	     	\State Update the position
	      \EndFor
	   \If{ condition met}
	   \State  Return the best global location as the global optimum
	   \EndIf
	   
		\EndFor
	\end{algorithmic}
	\label{tab: PSO_general}
\end{algorithm} 

In the literature,  some studies deployed PSO as an effective tool for solving large-scale optimization problems,  including optimal allocation of electric vehicle charging station and distributed renewable resource in power distribution networks    \cite{PSO2017app},  designing  power system stabilizers    \cite{PSO2002app},   distribution state estimation    \cite{PSO2001app},  and reactive power control    \cite{PSO2000app}. 
\subsection{Ant colony optimization (ACO)}
Ant colony optimization is another popular evolutionary algorithms which is presented in 1999 by Dorigo,  Marco and Di Caro    \cite{dorigo1999ant}, and Socha and Dorigo introduced continues domain of it    \cite{socha2008ant}. Basically,  ACO is one of stochastic search processes. Once Ants explored a new food source,  they try to lay some pheromone to mark the way which leads to the food. The pheromone is a chemical odorous material which is produced and used by ants to communicate with other ants in an indirect way. Each ant tries to produce it and lays it on their way. So Others can follow the odorous to seek for the food,  meanwhile they also produce the same amount of pheromone. On the other hand,   Further,  as we inspired natural behavior,  this chemical material is susceptible to be evaporate. Thus,  the amount of pheromone on specific path will increase by keeping ants on the same path,  However,  each iteration we have reduction which this amount have negative affect the total amount of pheromone on a particular path. In other words,  if any ants do not select the path used to be chosen,  then the path would disappear.  Algorithm \ref{tab: ACO_general} presents an overall procedure of ACO for feature selection.
\begin{algorithm}[H]
	\caption{Implementation of ACO algorithm for feature selection    \cite{kabir2012new} }
	\begin{algorithmic}[1]
   
		\Require  S=\{$x_0$, $x_1$, $x_2$, ..., $x_n$\} ,  K\} $\ge$ 1,   $\eta$ and $\tau$,   $ t=0 $,  $ best_{solution}$ =\O .
    	\Ensure $Best_{solution} :  An\ optimal\ subset\ of\ features\ ($F$)$ , $F=\{x_0, x_1, x_2, ..., x_m\}$ ,   m$\le$ n , $({\forall f_i \in F})\in S$ ,  $ F_{length} \le  S_{length}$.
		
    	\State {Call fitness function to calculate the fitness of each feature}
    	
    	\For {t=0  $\cdots$ $max_{Iteration}$}
    	\State {Generate K ants}
    	 \For {each ant $ \in Ants (K)$}
		  \State  {Generate\ a\ subset\ of\ features\ }
		  \State  call fitness function to evaluate the generated subset
		  \State Update the best local and global optimum
		  \If{ condition met}
	      \State  Return the best global location as the global optimum
	      \State {\textbf{else}}
	       	\State Update the $\eta$ and $\tau$
	   \EndIf
	   \EndFor
		\EndFor
	\end{algorithmic}
	\label{tab: ACO_general}
\end{algorithm} 

\subsection{Grey wolf optimizer (GWO)}
Grey Wolf Optimizer (GWO) is pretty new evolutionary algorithm which has been presented not sooner than 2014 which primary works based on the concept of grey wolf society    \cite{mirjalili2014GWOproposed}. Mirjalili and \emph{et al} claimed that    \cite{mirjalili2014GWOproposed} GWO outperforms other evolutionary algorithms for solving large-scale engineering and science problems. The GWO algorithm inspired by the natural mechanism of animals. The most common behavior which almost wild animal inherited normally are their attitude to have a kingdom,  rule others and  having the same hunting mechanism. It solves the science problems through the following steps: 

-First of all,  it searches for some animal as prey. In other words,  it tries to explore the area (food source);

-Then,  it  surrounds the possible prey(s) by exploitation,  doing local search to find the border of sample space;

-Finally,  it attacks the prey,  doing local search to find the best value within a new area. 'A' stands for the most important parameter in GWO and adjusts the step size towards the prey. Thus,  'A' has positive impact on convergence of this algorithm to the  global optimum  by tuning step size which influences both exploitation and exploration. 
However,  GWO still suffers from stalling in local minimum,  So initializing the parameter 'A' with a proper value helps it to prevent from stopping in local minimum.

\begin{algorithm}[H]
	\caption{Implementation of GWO algorithm for feature selection }
	\begin{algorithmic}[1]
   
		\Require  $S=\{x_0,  x_1, x_2, ..., x_n\}$,  $X_{i}=(i=1, 2, ...n)$,  $ A$, t=0 , $\alpha$,  $C$,  $ max_{iteration} \ge 0$,  $Best_{solution} \le  S_{length}$.
		
	\Ensure $Best_{solution} : $ $An $ $optimal $ $subset $ $of $$features$ $($F$)$ , $F=\{x_0, x_1, x_2, ..., x_m\}$ ,   m$\le$ n , $({\forall f_i \in F})\in S$ ,  $ F_{length} \le  S_{length}$.
    	
    	\State {Call fitness function for each search agent to evaluate the whole food source (S)  (primary evaluation of each food)}
    	\State { $X_{\alpha}$ ={ the best search agent}}
    	\State  {$X_{\beta} $={ the second best search agent}}
    	\State  {$X_{\delta} $={ the third best search agent}}
    	
		\For {t=0 $\cdots$ $max_{iteration}$}
		    \For {each search agent}
    		\State {$Update the best position of current search agent using \overrightarrow {X_{t+1}}=\frac{  \overrightarrow{X_1}+  \overrightarrow{X_2}+ \overrightarrow{X_3}}{3}$}.
    		
		     \EndFor
		     
		\State {Update $\alpha$ ,  A and C.}
		
		\State {Call fitness function to calculate the fitness of each search agent }
		 \State {Update $X_{\omega}$,  $X_{\beta}$,  $X_{\delta}$}
		
    	   \If{ any solution obtained the best score}
    	   \State   {Update the $Best_{solution}$ }
    	   \EndIf
    	   
		\EndFor
	\end{algorithmic}
	\label{tab: GWO_general}
\end{algorithm}

\subsection{Coyote optimization algorithm (COA)}
Coyote Optimization Algorithm (COA) is another yet important population-based meta-heuristic algorithms which have been inspired from the Canis latrans species and natural coyotes' behaviour. COA has a very certain procedure that works based on the way how these animals approaching other animals (preys) for catching them. Thus,  COA seems to be one particular type of Grey Wolf Optimizer (GWO) as COA just does the third step of GWO. COA is presented recently in    \cite{pierezan2018coyote} by Pierezan and Coelho in 2018 to solve large-scale optimization problems. Algorithm \ref{tab: COA_general} presents a general overview of COA for feature selection.

\begin{algorithm}[H]
	\caption{Implementation of COA algorithm for feature selection    \cite{pierezan2018coyote} }
	\begin{algorithmic}[1]
   
		\Require $S=\{x_0,  x_1, x_2, ..., x_n\}$ ${ which\ consists\ of\ } N_p \in N^* and N_c \in N^* {are\ initialized\ using\ } soc_{c,  j}^{p, t}=lower_{Bound j}+v_{j}\cdot(upper_{Bound j}-lower_{Bound j})$, t=0 ,  $max_{iteration}\ge 0$, 
		 $Best_{cayotes} = \emptyset$.
		
		\Ensure $Best_{cayotes} :  An\ optimal\ subset\ of\ features\ ($F$)$ , $F=\{x_0, x_1, x_2, ..., x_m\}$ ,   m$\le$ n , $({\forall f_i \in F})\in S$ ,  $ F_{length} \le  S_{length}$.
    	
    	\State {Call\ fitness\ function\ to\ calculate\ the\ coyote's\ fitness\ using: \ }
    	\State {$fit_{c}^{p, t}=f(soc_{c}^{p, t})$ }

		\For {t=0 $\cdots$ $max_{iteration}$}
		
		    \State {$ alpha^{p,  t}=\{soc_{c}^{p,  t}\vert arg_{c=\{1, 2, \ldots,  N_{c}\}}minf(soc_{c}^{p,  t})\} $}
		    	
		    \State  {Calculate the social tendency of the pack based on $N_c$ as follows: }
		    
		    \If{ $N_c$ \ is \ odd\ }
		    
		   \State{ $cult_{j}^{p, t}= \qquad\quad O_{\frac{(N_{c}+1)}{2}, j}^{p, t}$}
		    
		    \State {\textbf{else : }}
		    
		    \State { $cult_{j}^{p,  t}=\frac{O_{\frac{N_{c}}{2},  j}^{p,  t}+O_{(\frac{N_{c}}{2}+1),  j}^{p,  t}}{2}$  }
		     
		    \EndIf
		 
		    \For {each c coyotes in the p pack}
    		\State {Update \ the\ social\ condition\ using: }
    		\State{ ${\_}soc_{c}^{p,  t}=soc_{c}^{p,  t}+r_{1}\cdot\delta_{1}+r_{2}\cdot\delta_{2}$}
    		
    		\State{Examine\ the\ new\ social\ condition\ using: }
    		\State{$new{\_}fit_{c}^{p,  t}=f(new{\_}soc_{c}^{p,  t})$ }
    		
		    \State{ update\ food\ source\ with\ respect\ to\ better\ fitness\ using: }
		    \State {$soc_{c}^{p,  t+1}= new{\_}soc_{c}^{p,  t}$}
		    
		     \EndFor
		     \State {Birth\ and\ death\ using: }
		     \State {$pup_{j}^{p,  t}=\begin{cases} soc_{r_{1},  j}^{p,  t},  &\qquad\ rnd_{j} < P_{s}\ or\ j=j_{1}\\ soc_{r_{2},  j}^{p,  t},  &rnd_{j}\geq P_{s}+P_{a}\ or\ j=j_{2}\\ \quad\ R_{j},  &\qquad\qquad\qquad\ otherwise\end{cases}$}

    	   \State{$Transition\ between N_c and N_p\ packs\ using\ P_{e}=0.005\cdot N_{c}^{2}$}
    	   
    	   \State{Update\ the\ coyotes'\ information\ with\ respect\ to\ the\ age}
    	   
    	   \If{ stop condition met}
	             \State {$ Return\ the\ Best_{coyotes}$}
	             
	       \EndIf
	       
		\EndFor
	\end{algorithmic}
	\label{tab: COA_general}
\end{algorithm}

\subsection{Other optimization algorithms}
Meng \emph{et al}    \cite{meng2014new} proposed chicken swarm optimization algorithm (CSO) in 2014. Algorithm \ref{tab: CSO_general} presents well-structured pseudocode of CSo for optimized feature selection. Based on performance of CSO,  researchers have successfully solved and optimized engineering and science problems,  Directional Reader Antennas Optimization    \cite{shi2018optimizing}, Community detection in social networks   \cite{ahmed2016adaptive}, parameters Optimization of a fuzzy logic system    \cite{meng2017dempster}.
\begin{algorithm}[H]
	\caption{Implementation of CSO algorithm for feature selection  }
	\begin{algorithmic}[1]
   
		\Require  $S=\{x_0,  x_1, x_2, ..., x_n\}$,  $ N_p \in N^*$ ,  $N_c \in N^* {are\ done\ using\ } soc_{c, j}^{p, t}=lower_{Bound j}+v_{j}\cdot(upper_{Bound j}-lower_{Bound j})$ ,  t=0 , $rooster_{ratio}$, $chicks_{ratio}$, $hens_{ratio}$,  $food_{position}\ C$,  $Random_{value}$,  $min_{iteration}$, $max_{iteration}$, $chickenSwarm_{size}$ .
		
	\Ensure $Best_{solution} :  An optimal subset of features ($F$)$ , $F=\{x_0, x_1, x_2, ..., x_m\}$ ,   m$\le$ n , $({\forall f_i \in F})\in S$ ,  $ F_{length} \le  S_{length}$.
   
   		\For {t=0 $\cdots$ $max_{iteration}$}
		    \State {call fitness function to compute the fitness using chicken}
		    	
		    \If{fitness \ of \ chicken\ ==$ best_{fitness} $}
		    \State{$Update\ the\ Random_{value}$} 
		    \State{$Update\ the\ rooster\ position$}
		    \EndIf
		    \If{fitness \ of \ chicken\ ==$ worst_{fitness}$ }
		 	    \State{Update the chicks position}
		    \EndIf
		    \If{fitness \ of \ chicken\ != $worst_{fitness}$ and 
		      fitness \ of \ chicken\ != $best_{fitness}$ }
		 	    \State{$Update\ the\ Random_{value}$}
		        \State{$Update\ the\ hens\ position$}
		    \EndIf
		     
		    \State{$Update\ chicken\ position$}
		    
		   \If{t==$chickenSwarm_{size}$}
		     
	          \State  Return the best position as the global optimum
    	    \EndIf
		\EndFor
	\end{algorithmic}
	\label{tab: CSO_general}
\end{algorithm} 
 Li \emph{et al} introduced fish swarm algorithm (FSA) which is another population-based ( or swarm-based) evolutionary algorithm    \cite{li2002optimizing}. FSA inspired from the behaviors of fish school. Algorithm \ref{tab: FSA_general} shows the process of feature selection using FSA. research studies have applied FSA to optimize their solution such as  neighborhood feature selection    \cite{chen2017neighborhood},  multi-modal benchmark functions solver    \cite{rahman2019artificial}.

\begin{algorithm}[H]
	\caption{Implementation of FSA algorithm for feature selection    \cite{pierezan2018coyote} }
	\begin{algorithmic}[1]
   		\Require  $S=\{x_0,  x_1, x_2, ..., x_n\}$,  t=0 , $ max_{iteration} \ge 0$,  $R_{min}$,  $L_{min}$,  $_{\gamma_B}(D)=\frac{|POS_B(D)_\gamma|}{|U|}$, 
		$Best_{cayotes} = \emptyset$.
		
		\Ensure { $best_{cayotes}$:  An\ optimal\ subset\ of\ features\ (F)\ } , $R_{min}=$ $F=\{x_0, x_1, x_2, ..., x_m\}$  ,  m$\le$ n , $({\forall f_i \in F})\in S$ ,  $ F_{length} \le  S_{length}$.
    	
    \State {$R_{min}$=C , $L_{min}$=C}
    	
		\For {t=0 $\cdots$ $max_{iteration}$}

		    \State  {generate total fish (Fish)}
		    \For {each fish $K \in Fish$}
		    
		    \State {$R_K$=\O , $L_K$=0}
		    \State{$Choose\ a\ feature\ \alpha_k \in C (randomly)$}
		    
		    \State {$Update{R_K,  L_K} by {R_K\bigcup \alpha_K} and |R_K|,  respectively$}
		    
		    \EndFor
		    
		    \For{each fish$ K \in Fish$}
		    
		    \State{$R_s=Search(R_k)$}
		    
		    \State{$R_{\omega}=Swarm(R_k)$}
		    
		    \State{$R_f=Follow(R_k)$}
		    
		     \State Update{$R_K$,  $L_K$} by seeking for the $max_{fitness}$ through ($R_k$, $R_\omega$, $R_f$)
		     
		    \If{$_{\gamma R_k}(D)_\delta ==_{\gamma C}(D)_\delta$}
		    
		    \State{$The\ fish_K\ obtained\ a\ local\ reduction\ and\ break$ }
		    
		    \EndIf
		    \If{$_{\gamma R_k}(D)_\delta ==_{\gamma C}(D)_\delta and L_K \le L_{min}$}
		    
		    \State {$update{R_{min}}, L_{min} by  R_K and L_K,  respectively$}
		    		     
		    \EndIf
		    \EndFor
		 
    	   \If{ stop condition met}
	             \State  {$Return\ the\ R_{min}, L_{min}$}
	             
	       \EndIf
	       
		\EndFor
	\end{algorithmic}
	\label{tab: FSA_general}
\end{algorithm} 

\section{Conclusion}

Both dimension reduction by generating new dimension of features and feature selection by eliminating irrelevant and redundant features take care of missing values and classify supervised / unsupervised data sets; all of these operations come together to solve emerging challenging Np-hard problems in engineering and the sciences. A large number of data sets,  particularly Big Data,  are available to work on. The main problem,  here,  concerns their features and dimensionality,  the curse of dimensionality (CoD),  which causes yet another important problem,  high time complexity. In this chapter,  we addressed these problems and professional approaches using advanced machine learning algorithms. The studies prove that applying nature-inspired algorithms,  together with machine learning techniques,  enabled researchers' attempts to solve the CoD problem,  which yields a proper running time with a lowest time complexity. It is noteworthy that evolutionary algorithms are non-dependent domain specific,  which provides an  optimized environment for researchers who want to solve their problems or optimize their approaches. In this chapter,  we have explored  evolutionary algorithms and their applications in solving large scale optimization problems,  especially the feature extraction process for data analytics. This chapter provides insightful information for researchers who are seeking for the application of evolutionary algorithms for engineering,   optimization,  and data science. Having said this,  in    \cite{ch2_farid},  we address the emerging problem,  CoD,  and an evolutionary-based solution is presented to solve it. We discuss the feature extraction optimization process in detail,  leveraging feature extraction and evolutionary algorithms. Then,  we provide detailed and practical examples of applying evolutionary algorithms with a wide variety of domains. we also classify all research studies  based on the most common challenging issues such as stego image classification,  network anomalies detection,  network traffics classification,  sentiment analysis and supervised benchmark classification.
.
\bibliographystyle{unsrt}
\bibliography{bib.bib}
\end{document}